\documentclass[conference]{IEEEtran}
\IEEEoverridecommandlockouts
\usepackage{cite}
\usepackage{amsmath,amssymb,amsfonts}
\usepackage{graphicx}
\usepackage{textcomp}
\usepackage{xcolor}
\usepackage{subcaption}
\usepackage{geometry}[left=54pt,right=37pt,top=105pt,bottom=54pt]
\usepackage{changepage}
\usepackage{float}
\usepackage{listings}
\usepackage{xcolor} 
\usepackage{algorithm}       
\usepackage{algpseudocode}     

\lstdefinelanguage{json}{
    basicstyle=\ttfamily\footnotesize,
    breaklines=true,
    showstringspaces=false,
    string=[s]{"}{"},
    stringstyle=\color{red},
    comment=[l]{//},
    morecomment=[s]{/*}{*/},
    commentstyle=\color{gray}\ttfamily,
    keywordstyle=\color{blue}\bfseries
}

\newgeometry{left=54pt,right=37pt,top=122pt,bottom=54pt}
\def\BibTeX{{\rm B\kern-.05em{\sc i\kern-.025em b}\kern-.08em
    T\kern-.1667em\lower.7ex\hbox{E}\kern-.125emX}}

\begin{document}


\title{VirtualXAI: A User-Centric Framework for Explainability Assessment Leveraging GPT-Generated Personas \\
}

\author
{\IEEEauthorblockN{1\textsuperscript{st} Georgios Makridis}
\IEEEauthorblockA{\textit{Department of Digital Systems} \\
\textit{University of Piraeus}\\
Piraeus, Greece \\
gmakridis@unipi.gr}
\and
\IEEEauthorblockN{2\textsuperscript{nd} Vasileios Koukos}
\IEEEauthorblockA{\textit{Department of Digital Systems} \\
\textit{University of Piraeus}\\
Piraeus, Greece \\
vkoukos@unipi.gr}
\and
\IEEEauthorblockN{3\textsuperscript{rd} Georgios Fatouros}
\IEEEauthorblockA{\textit{Department of Digital Systems} \\
\textit{University of Piraeus}\\
Piraeus, Greece \\
gfatouros@unipi.gr}
\and
\IEEEauthorblockN{4\textsuperscript{th} Dimosthenis Kyriazis}
\IEEEauthorblockA{\textit{Department of Digital Systems} \\
\textit{University of Piraeus}\\
Piraeus, Greece \\
dimos@unipi.gr}
}

\maketitle

\begin{abstract}
In today’s data-driven era, computational systems generate vast amounts of data that drive the digital transformation of industries, where Artificial Intelligence (AI) plays a key role. Currently, the demand for eXplainable AI (XAI) has increased to enhance the interpretability, transparency, and trustworthiness of AI models. However, evaluating XAI methods remains challenging: existing evaluation frameworks typically focus on quantitative properties such as fidelity, consistency, and stability without taking into account qualitative characteristics such as satisfaction and interpretability. In addition, practitioners face a lack of guidance in selecting appropriate datasets, AI models, and XAI methods —a major hurdle in human–AI collaboration. To address these gaps, we propose a framework that integrates quantitative benchmarking with qualitative user assessments through virtual personas based on the “Anthology” of backstories of the Large Language Model (LLM). Our framework also incorporates a content-based recommender system that leverages dataset-specific characteristics to match new input data with a repository of benchmarked datasets. This yields an estimated XAI score and provides tailored recommendations for both the optimal AI model and the XAI method for a given scenario.

\end{abstract}


\section{Introduction}
In today’s data-driven environment, advanced computational systems generate vast amounts of data that fuel the digital transformation of industries. This phenomenon has led us into the era of Industry 4.0 \cite{makridis2020predictive}, where Artificial Intelligence (AI) plays a key role in driving innovation across sectors. At the same time, the demand for eXplainable Artificial Intelligence (XAI) has risen to improve the interpretability, transparency, and trustworthiness of AI models \cite{saarela2024recent, arrieta2020explainable}. This is particularly important in safety critical industries, such as \cite{makridis2023deep} and regulated industries (\cite{kotios2022personalized, fatouros2022deepvar}), where explainability and transparency are essential to gain the trust of stakeholders and ensure compliance with legal and ethical requirements.

Recent studies have shown that the most prominent XAI techniques are SHAP \cite{lundberg2017unified} and LIME \cite{ribeiro2016should}. Although SHAP is well-regarded for its stability and mathematical foundations, LIME is appreciated for its model-agnostic properties despite some noted instability \cite{saarela2024recent}. However, evaluating remains a challenging task, with most approaches relying on quantitativequantitative anecdotald expert opinion \cite{nauta2023xai}. Existing evaluation frameworks have typically focused on properties such as fidelity, consistency, stability, and certainty \cite{ribeiro2016should, lundberg2017unified}, yet these approaches are often tailored to specific methods \cite{lage2019evaluation}. Furthermore, \cite{samek2017explainable , doshi2017towards} have emphasized the necessity of interpretable models and have outlined the challenges in objectively measuring the quality of explanation. Moreover, the lack of knowledge to select the appropriate AI models and XAI methods has been highlighted as a major hurdle in human-AI collaboration \cite{makridis2023unified}. \newline
Large Language Models (LLMs) have recently emerged as powerful tools generating human-readable explanations and bridging the gap between technical algorithms and domain understanding \cite{fatouros2025marketsenseai}. Their natural language processing capabilities offer significant potential for improving the accessibility and interpretability of complex AI systems, particularly in domains where human judgment and regulatory compliance are paramount or in specialized fields with technical terminology \cite{fatouros2023transforming}. To address this gap, we propose an XAI scoring framework that integrates quantitative benchmarking with qualitative user assessments through virtual personas based on the GPT-4-mini generated Anthology of backstories, following the results of \cite{moon2024virtual}. In addition, our framework incorporates a content-based recommendation system that uses dataset-specific characteristics to match input data with a repository of benchmarked datasets, thus estimating an XAI score and providing tailored recommendations for both AI and XAI methods.

The contributions of this paper are threefold:
\begin{enumerate}
    \item Development of a XAI Scoring Framework for tabular data that integrates fidelity, simplicity, stability, and accuracy metrics.
    \item Introduction of an LLM-based qualitative assessment methodology to capture user-centric qualitative assessment.
    \item Creation of a content-based recommender system to assist users in selecting datasets, AI models, and XAI methods by matching dataset characteristics to historical benchmarks.
\end{enumerate}

The remainder of the paper is organized as follows. Section 2 provides a review of existing XAI methods and their evaluation frameworks. In Section 3, we introduce our proposed explainability framework, detailing its mathematical formulation and the integration of various XAI properties. Section 4 presents our experimental evaluation of the explainability metric across diverse datasets. Finally, Section 5 offers conclusions and recommendations for future work.

\section{Related Work}

In recent years, several off-the-shelf frameworks and toolkits have been proposed to facilitate the application of explainability methods to AI models, such as AI Explainability 360 \cite{arya2021ai} and InterpretML \cite{nori2019interpretml}. These frameworks offer guidelines and implementations for various explainability techniques. 

Moreover, several surveys and frameworks have been developed to evaluate black-box models \cite{guidotti2018survey} and empirically assess the impact of explanation quality on end-user trust and performance \cite{poursabzi2021manipulating}. For example, \cite{stassin2023experimental} conducted an experimental investigation comparing 14 different metrics across nine state-of-the-art XAI methods. Their findings highlight correlations among certain metrics, and notable limitations in the reliability of these metrics.

\cite{bommer2023guide} introduced a guide to evaluate and rank XAI methods. Their work assesses explanation properties such as robustness, faithfulness, randomization, complexity, and localization across multiple XAI techniques applied to climate data. Similarly, the OpenHEXAI framework \cite{ma2024openhexai} offers an open source platform designed for the human-centered evaluation of explainable machine learning. This framework integrates benchmark datasets, pre-trained models, and post hoc explanation methods. Another contribution is the development of the XAI Experience Quality Scale (XEQ) by \cite{wijekoon2024xai}. Grounded in psychometric theory, the XEQ evaluates user-centered aspects of XAI experiences by measuring dimensions such as learning, utility, fulfillment, and engagement.

In \cite{narayanan2018humans}, a human-centered approach is presented that explores how users understand explanations generated by machine learning systems. \cite{beede2020human} focuses on a specific application by evaluating the explanations provided by a deep learning system for diabetic retinopathy detection. Their user study with medical professionals assessed the impact of explanations on performance and trust, demonstrating the practical implications of effective explainability in high-stakes decision-making contexts. Similarly, \cite{pruthi2022evaluating} emphasizes not only the generation of high-quality explanations but also their effective communication to users, proposing the quantification of transmitted information using methods from Information Theory. Alongside these efforts, \cite{litman2023measures} introduces user-centered metrics—including user satisfaction, mental models, curiosity, trust, and the performance of human-AI collaborations—to present a more comprehensive picture of explainability. One step forward towards personalized XAI was made by ehe x-[plAIn] framework. It demonstrates how domain-specific LLMs can democratize XAI accessibility. Developed using ChatGPT Builder, this system adapts explanations to audience expertise levels \cite{mavrepis2024xai}.

In contrary to the emerging need and applications of XAI, quantifying explainability remains a complex task due to its multifaceted nature and the diverse range of stakeholders involved. The inherent subjectivity —where different individuals may have, poses significant challenges in designing a universally applicable explainability metric. One potential solution is to incorporate user feedback into the evaluation process, allowing the metric to adapt to individual preferences via active learning \cite{holzinger2019interactive}. Moreover, trade-offs often exist between different aspects of explainability, such as simplicity, fidelity, and coverage; a highly accurate and detailed explanation might be more complex and harder for users to understand, whereas a simpler explanation might sacrifice some fidelity for improved comprehensibility \cite{gilpin2018explaining}.

These newer perspectives underscore the need for a holistic evaluation of XAI techniques that encompasses both technical and user-centric aspects. Our work builds on these insights by proposing a methodology that benchmarks XAI techniques against a wide range of criteria, integrating these multifaceted approaches.

\section{Methodology}

\begin{figure*}[h!]
\centering
\includegraphics[width=0.7\textwidth]{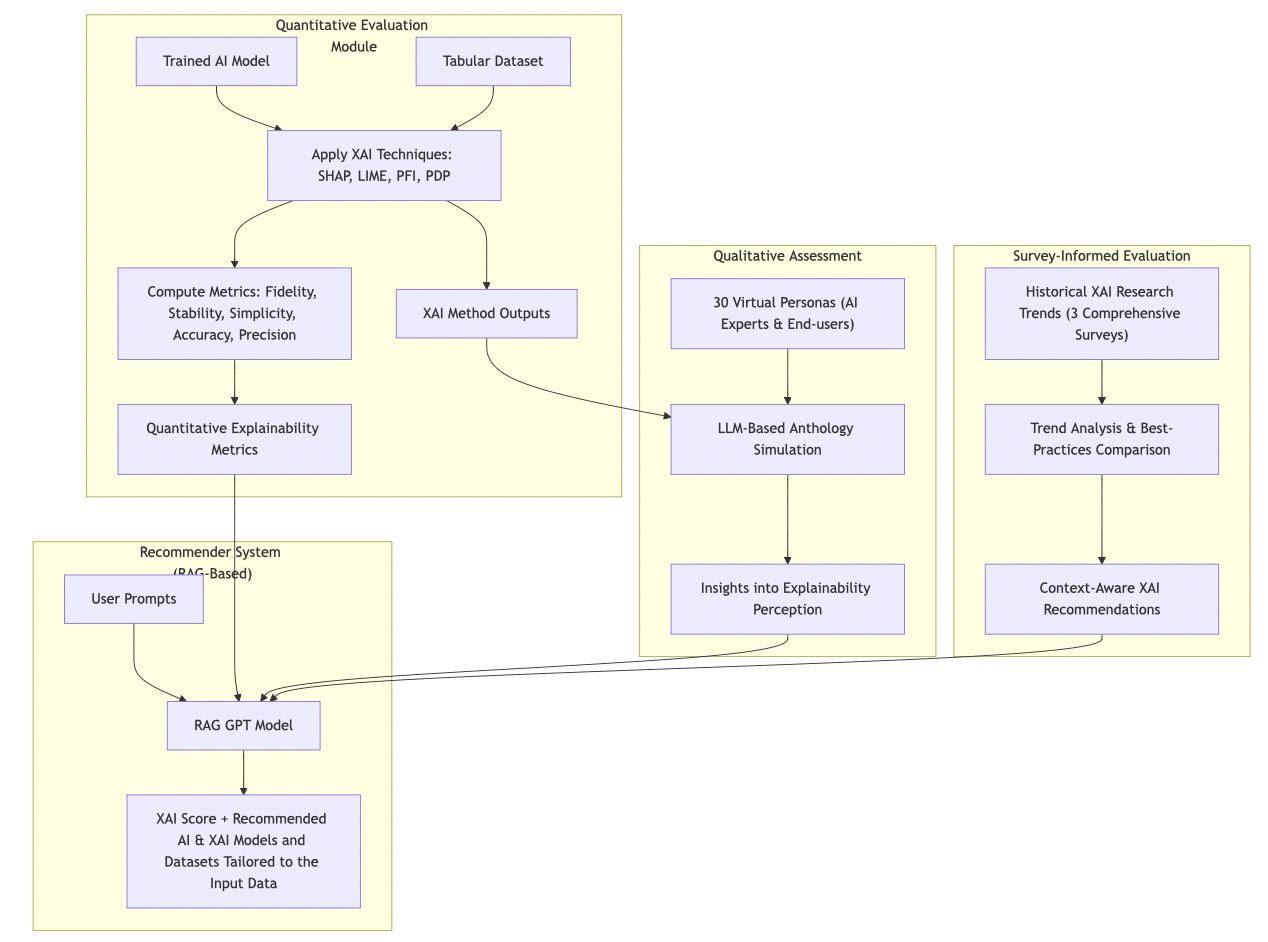}
\caption{Integrated system architecture combining quantitative evaluation, qualitative assessment, and survey-informed insights. The recommender system uses these inputs to generate an XAI score and recommend AI and XAI methods based on the dataset’s characteristics and user preferences.}
\label{fig:arch}
\end{figure*}

\begin{figure}[h!]
\centering
\includegraphics[width=0.4\textwidth]{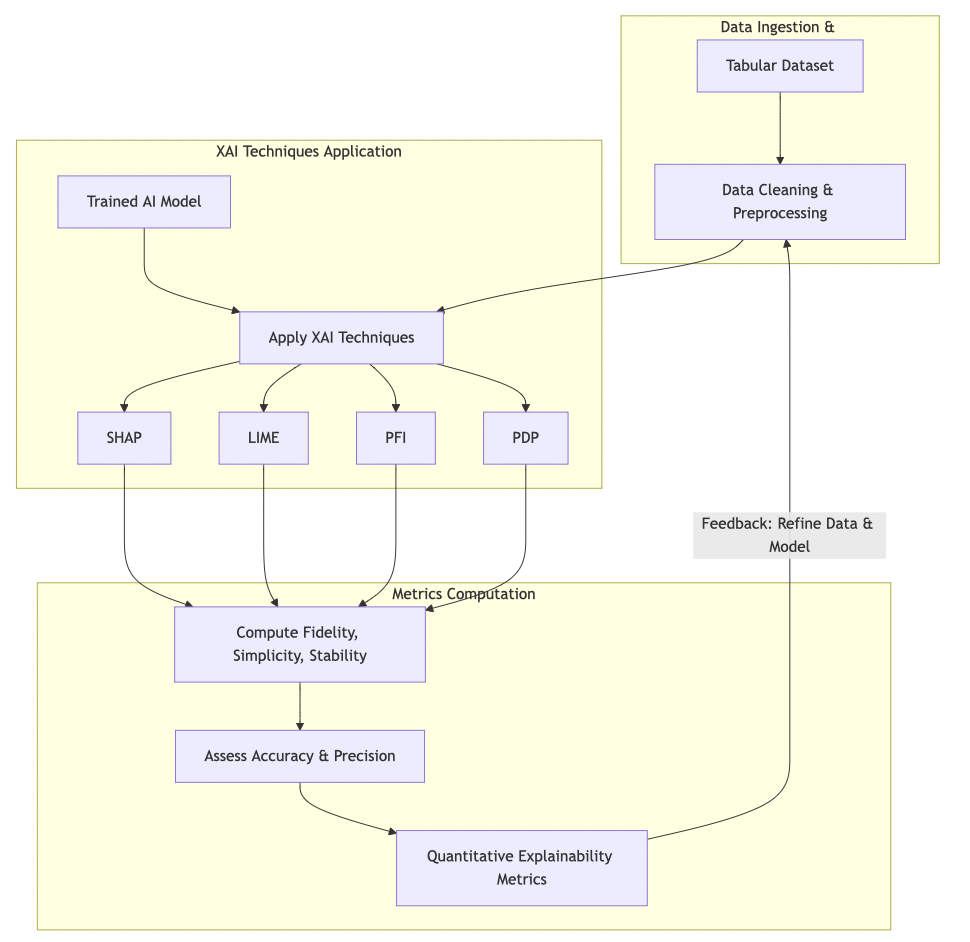}
\caption{High-level pipeline for the quantitative evaluation of XAI methods. Data is ingested and preprocessed, then a trained AI model is explained using SHAP, LIME, PFI, or PDP}
\label{fig:quant}
\end{figure}

\begin{figure}[h!]
\centering
\includegraphics[width=0.4\textwidth]{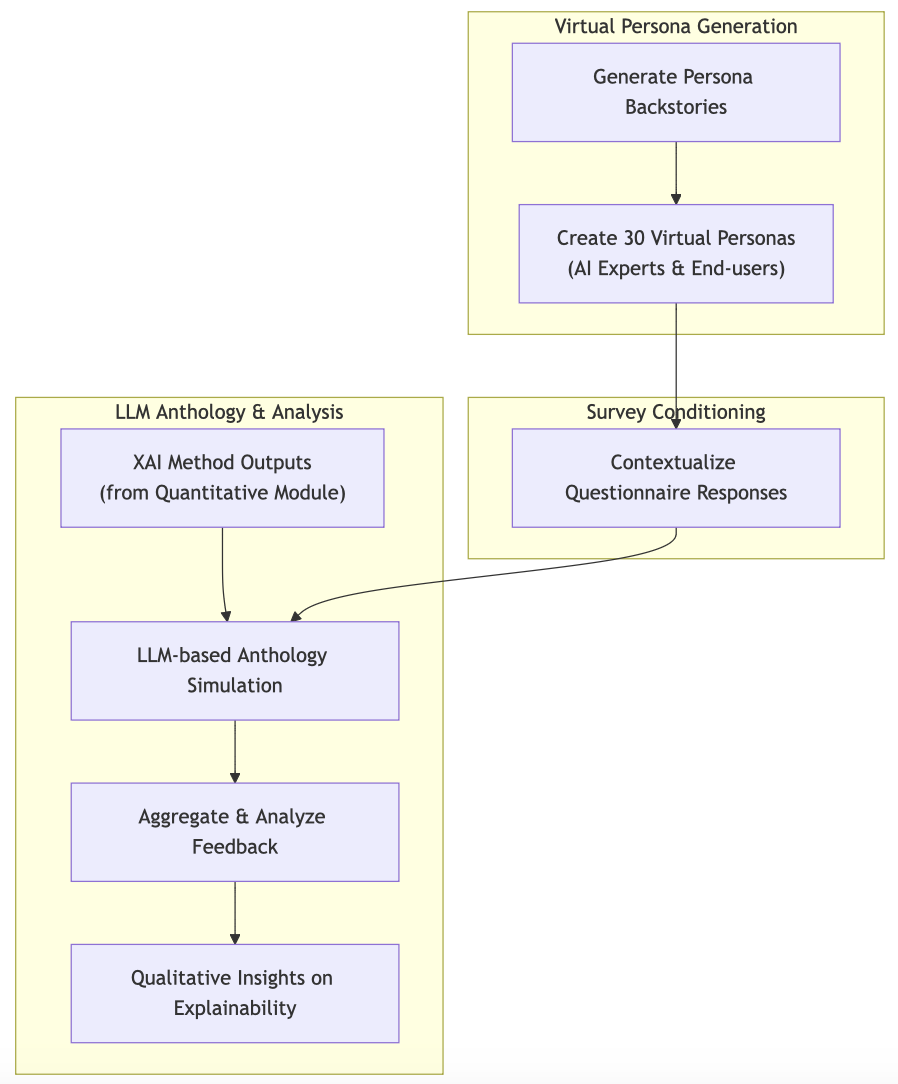}
\caption{Overview of the qualitative assessment approach. Virtual personas are generated, and their feedback on the XAI outputs is collected through structured surveys. }
\label{fig:qual}
\end{figure}

The approach is based on the XAI evaluation strategy presented in \cite{makridis2023unified} and further enhanced with qualitative assessment levergaing LLM generated virtual personal based on an anthology of backstories and insights from three surveys \cite{saarela2024recent} \cite{nauta2023anecdotal} \cite{nagahisarchoghaei2023empirical}. As depicted in Figure~\ref{fig:arch}, the system architecture comprises four primary components: (1) a quantitative evaluation module, (2) a qualitative assessment using virtual personas, (3) a survey-informed evaluation process, and (4) a content-based recommender system for generating an overall XAI score.

\subsection{System Architecture Overview}

\begin{enumerate} 
\item \textbf{Quantitative Evaluation}: Benchmarks XAI methods by measuring fidelity, stability, simplicity, and accuracy/precision. A high-level pipeline for this process is shown in Figure~\ref{fig:quant}. 
\item \textbf{Qualitative User Assessment}: Uses GPT-4o-mini to generate virtula personas based on an LLM-generatred Anthology of backstories. And these virtual personas respond to a tailored questionaure to capture user preferences and interpretability requirements. This approach, illustrated in Figure~\ref{fig:qual} (generates virtual personas and aggregates their feedback). 
\item \textbf{Survey-Informed Model Evaluation}: Integrates findings from large-scale surveys to reflect current best practices and trends across multiple domains. \item \textbf{Content-Based Recommender System}: Leverages dataset characteristics to estimate an XAI score and recommend proper AI and XAI methods. 
\end{enumerate}

\subsection{Quantitative Metrics for Benchmarking XAI Methods}
The quantitative evaluation focuses on benchmarking four widely used XAI techniques —SHAP, LIME, PFI, and PDP—against tabular datasets. Following the methodology described in \cite{makridis2023unified}, the key metrics for evaluation include:
\begin{itemize}
    \item \textbf{Fidelity}: Measures the degree to which an explanation accurately reflects the behavior of the underlying model.
    \item \textbf{Simplicity}: Assesses the interpretability of the explanation by quantifying its complexity.
    \item \textbf{Stability}: Evaluates the consistency of explanations when minor perturbations are applied to the input data.
    \item \textbf{Accuracy and Precision}: Quantifies the trade-off between the performance of the model and the quality of its explanations.
\end{itemize}
For each XAI method, these metrics are computed using a reproducible and robust procedure as detailed in \cite{makridis2023unified}.

\subsection{Qualitative Assessment}

The qualitative assessment via virtual personas addresses several challenges inherent in traditional user satisfaction studies. Recruiting real end users can be time-consuming, costly, and prone to biases due to inconsistent participation. By generating virtual personas with carefully designed demographic and professional profiles —as outlined in \cite{moon2024virtual}— our process simulates a broad range of real-world perspectives in a consistent and scalable manner. Figure~\ref{fig:qual} lustrates the process that involves:
\begin{itemize}
    \item \textbf{Virtual Personas Generation}: Using GPT-4o-mini (via openai api) we generated 1000 backstories by applying open quesitons. Then we chose in a balance way 100 out of 1000 backstories and based on them the virtual personas are generated using LLMs, each with a unique demographic and professional profile, including factors such as age, profession, level of AI expertise, and specific explainability preferences.
    \item \textbf{Survey Conditioning}: Each persona's profile conditions a structured questionnaire aimed at evaluating XAI methods. 
    \item \textbf{Feedback Analysis}: The responses from these virtual personas are aggregated and analyzed to user satisfaction and preferences, providing qualitative insights into the interpretability of model explanations.
\end{itemize}

\subsection{Comprehensive Review of Sector-Specific XAI Applications}

To ensure our framework is aligned with current research, we integrate insights from three comprehensive surveys on XAI methodologies, ensuring that our recommendations reflect domain-specific trends and real-world applicability. These surveys provide: 

\begin{itemize} 
\item Detailed breakdowns of XAI adoption across industries such as healthcare and finance. 
\item Comparative analyses of XAI methods, including their strengths and limitations across different domains. 
\item Assessments of trends in AI model usage, evaluation strategies, and user preferences. 
\end{itemize} 

For illustration purposes, we include example JSON listings that capture key empirical data. Listing~\ref{lst:xai_explanation_frequency} shows the frequency of various XAI explanation techniques, and Listing~\ref{lst:xai_models_per_domain} outlines the most frequent XAI models employed across different domains. It is important to note that these listings are examples and not the final versions.

\begin{lstlisting}[language=json, caption={XAI Explanation Techniques Frequency}, label={lst:xai_explanation_frequency}]
{
  "xai_explanation_techniques_frequency": {
    "SHAP": 175,
    "LIME": 125,
    "Grad-CAM": 75,
    "Decision_Tree": 35,
    "Permutation_Importance": 25,
    "Integrated_Gradients": 20,
    "SmoothGrad": 15,
    "LRP": 10,
    "PDP": 8,
    "Grad-CAM++": 7,
    "Anchors": 6,
    "Logistic_Regression": 5,
    "ALE": 5,
    "CAM": 5,
    "RISE": 5
  }
}
\end{lstlisting}

\begin{lstlisting}[language=json, caption={Most Frequent XAI Models in Different Domains}, label={lst:xai_models_per_domain}]
{
  "most_frequent_xai_models_per_domain": {
    "healthcare": ["SHAP", "Grad-CAM"],
    "finance": ["SHAP", "LIME"],
    "cybersecurity": ["PFI", "SHAP"],
    "transportation": ["PDP", "Integrated Gradients"],
    "education": ["LIME", "Decision Trees"]
  }
}
\end{lstlisting}

\subsection{Development of a Content-Based Recommender System}

To assist end-users in selecting appropriate datasets, AI models, and XAI methods, we propose a content-based recommender system. This leverages the intrinsic characteristics of the input dataset, such as feature distribution, dimensionality, and domain. Its operation comprises two phases:

\begin{enumerate}
    \item \textbf{Dataset Matching}:  
    \begin{itemize}
        \item Feature Extraction: Features are extracted from the input dataset (e.g., statistical summaries, feature distributions, and domain indicators).
        \item Similarity Assessment: The input dataset is compared against a repository of previously benchmarked datasets using content-based filtering techniques to identify similar datasets.
    \end{itemize}
    \item \textbf{Recommendation Generation}:  
    \begin{itemize}
        \item XAI Score Estimation: Based on the performance metrics obtained from matched datasets, an estimated XAI score is computed for the target dataset.
        \item Method Recommendation: The system recommends optimal AI and XAI methods that have historically yielded high XAI scores on similar datasets.
    \end{itemize}
\end{enumerate}

\begin{algorithm}[ht]
\caption{Benchmarking Phase: Quantitative \& Qualitative Benchmarking and Information Extraction}\label{alg:training}
\begin{algorithmic}[1]
\Require 
  \begin{itemize}
      \item Benchmark datasets $\mathcal{D}$
      \item Trained AI models $\mathcal{M}$ (or training procedure for each $D\in\mathcal{D}$)
      \item XAI techniques $\mathcal{X} = \{\text{SHAP}, \text{LIME}, \text{PFI}, \text{PDP}\}$
      \item Number of virtual personas $P$
  \end{itemize}
\Ensure Repository $\mathcal{R}$ containing for each dataset: 
\begin{itemize}
    \item Quantitative metrics $Q_D$ (fidelity, simplicity, stability, accuracy, precision)
    \item Qualitative insights $Q_{qual,D}$
    \item Dataset characteristics $C_D$
\end{itemize}
\For {each dataset $D \in \mathcal{D}$}
    \State Preprocess dataset $D$ (e.g., data cleaning, normalization, encoding)
    \State Train AI model $M_D$ on $D$ (or use an existing model from $\mathcal{M}$)
    \For {each XAI technique $x \in \mathcal{X}$}
        \State Generate explanation $E_{x}$ for $M_D$ using $x$
        \State Compute quantitative metrics: 
            \begin{itemize}
                \item Fidelity: How accurately $E_{x}$ reflects $M_D$
                \item Simplicity: Complexity measure of $E_{x}$
                \item Stability: Consistency of $E_{x}$ under input perturbations
                \item Accuracy \& Precision: Trade-offs between model performance and explanation quality
            \end{itemize}
        \State Store metrics as $Q_{x,D}$
    \EndFor
    \State Aggregate quantitative metrics: $Q_D \gets \{ Q_{x,D} \mid x \in \mathcal{X} \}$
    \State Extract dataset characteristics $C_D$ (e.g., feature distributions, dimensionality, sparsity)
    \State \textbf{Qualitative Assessment:}
    \State Generate $P$ virtual personas with diverse backstories using GPT-4o-mini
    \State Condition a structured questionnaire with and apply it to $E = \{E_x \mid x\in\mathcal{X}\}$
    \State Aggregate the survey responses to derive qualitative insights $Q_{qual,D}$
    \State Store the tuple $(D, Q_D, Q_{qual,D}, C_D)$ in repository $\mathcal{R}$
\EndFor
\State \Return $\mathcal{R}$
\end{algorithmic}
\end{algorithm}

\begin{algorithm}[ht]
\caption{Inference Phase: Dataset Matching, XAI Score Estimation and Recommendation}\label{alg:inference}
\begin{algorithmic}[1]
\Require 
  \begin{itemize}
      \item User-uploaded dataset $D_u$
      \item Repository $\mathcal{R}$ containing tuples $(D, Q_D, Q_{qual,D}, C_D)$ from the training phase
  \end{itemize}
\Ensure 
  \begin{itemize}
      \item Estimated XAI score $S_{XAI}$ for $D_u$
      \item Recommendations for optimal AI model $M^*$ and XAI method $x^*$
  \end{itemize}
\State Preprocess the uploaded dataset $D_u$ (cleaning, normalization, encoding)
\State Extract dataset characteristics $C_u$ from $D_u$
\State Compute similarity between $C_u$ and each stored $C_D$ in $\mathcal{R}$ using a similarity metric
\State Identify the top $k$ matching datasets $\{D_{(1)},\dots,D_{(k)}\}$ based on similarity scores
\State Aggregate the corresponding quantitative metrics and qualitative insights from the matched datasets
\State Estimate the XAI score $S_{XAI}$ for $D_u$ using the aggregated metrics (e.g., via weighted averaging)
\State Determine the optimal AI model and XAI method recommendation $(M^*, x^*)$ based on historical performance on similar datasets
\State \Return $S_{XAI}$, $M^*$, and $x^*$
\end{algorithmic}
\end{algorithm}

\section{Results}
To assess the effectiveness of the proposed XAI Scoring Framework, we conducted a series of benchmarking experiments using a diverse set of tabular datasets and explainability techniques. 

\subsection{Datasets}
We utilized a diverse set of tabular datasets from the UCI repository \cite{asuncion2007uci}, each representing distinct domains and varying levels of feature complexity. All datasets underwent standardized preprocessing procedures, including categorical encoding and normalization, to ensure consistency. A Random Forest classifier was trained on each dataset, after which the selected XAI techniques were applied to interpret the model predictions. This approach allowed us to evaluate the performance of the explainability methods in a domain-agnostic manner.

\subsection{Dataset Distribution Across Domains}
\begin{figure}[htbp]
\centering
\includegraphics[width=0.48\textwidth]{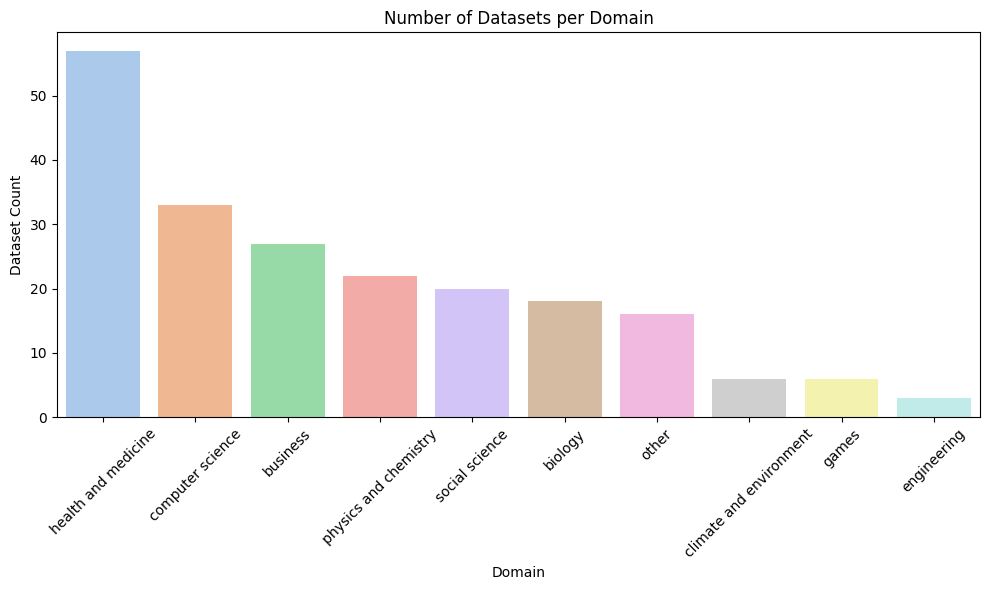}
\caption{Number of datasets per domain. The \emph{health and medicine} domain has the highest dataset count, reflecting a significant interest in clinical AI applications.}
\label{fig:datasets_domain}
\end{figure}

The utillized datasets span multiple domains, as illustrated in Figure~\ref{fig:datasets_domain}. Notably, the \emph{health and medicine} domain exhibits the largest number of datasets (over 50), followed by \emph{computer science}, \emph{business}, and \emph{physics and chemistry}. This distribution highlights the prominence of medical and clinical use cases in current AI research, a trend corroborated by our survey-based data. 

\subsection{Quantitative Benchmarking of XAI Methods}

\begin{figure}[htbp]
\centering
\includegraphics[width=0.48\textwidth]{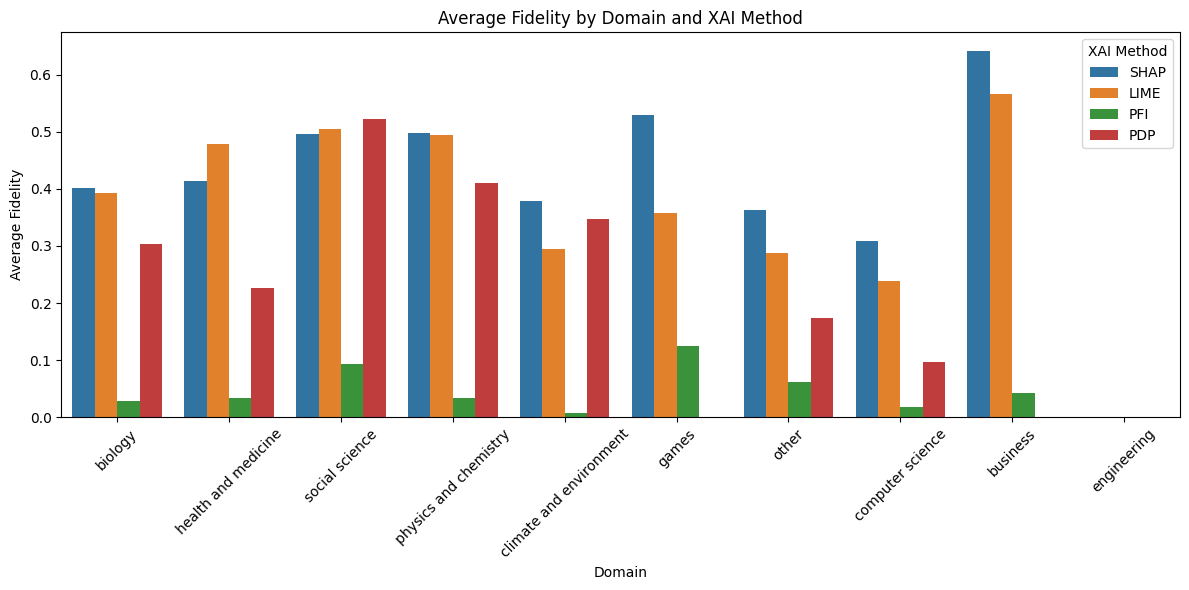}
\caption{Average fidelity by domain and XAI method. Higher bars indicate stronger alignment between the explanation and the model's predictions.}
\label{fig:avg_fidelity_domain}
\end{figure}

Figure \ref{fig:avg_fidelity_domain} presents the average fidelity of four XAI methods—SHAP, LIME, PFI, and PDP—across multiple domains. The results highlight a clear variation in method performance depending on the domain. In health and medicine, SHAP demonstrates consistently high fidelity, suggesting it aligns well with clinical data. Conversely, PDP exhibits notably higher fidelity in business applications, indicating it may be particularly effective for the kinds of features and relationships found in that domain. LIME and PFI maintain relatively moderate yet steady performance across most domains, with occasional spikes in areas such as biology or computer science. Overall, these findings underscore that no single XAI method dominates in every context, reinforcing the importance of domain-specific considerations when selecting an explainability technique.

Table~\ref{tab:xai_scores} provides an additional breakdown for selected datasets. For example, on a Heart Disease dataset within the \emph{health and medicine} domain, SHAP attains a fidelity of 0.82, while LIME exhibits a lower simplicity value (5.1), indicating more concise explanations. PFI excels in stability (0.93), underscoring its consistency under feature perturbations.

\begin{table}[htbp]
\centering
\begin{tabular}{lcccc}
\textbf{Dataset} & \textbf{Method} & \textbf{Fidelity} & \textbf{Simplicity} & \textbf{Stability} \\
Heart Disease & SHAP & 0.82 & 7.3 & 0.91 \\
              & LIME & 0.79 & 5.1 & 0.88 \\
              & PFI  & 0.76 & 4.6 & 0.93 \\
              & PDP  & 0.74 & 6.0 & 0.87 \\
Wine Quality  & SHAP & 0.85 & 6.8 & 0.89 \\
              & LIME & 0.78 & 5.3 & 0.85 \\
              & PFI  & 0.74 & 4.1 & 0.92 \\
              & PDP  & 0.71 & 5.7 & 0.84 \\
\end{tabular}
\caption{Sample of quantitative explainability scores for SHAP, LIME, PFI, and PDP. Fidelity measures alignment with the model, Simplicity indicates fewer features (lower is better), and Stability measures consistency.}
\label{tab:xai_scores}
\end{table}

\subsection{Qualitative User Ratings and Interpretability}
\begin{figure}[htbp]
\centering
\includegraphics[width=0.48\textwidth]{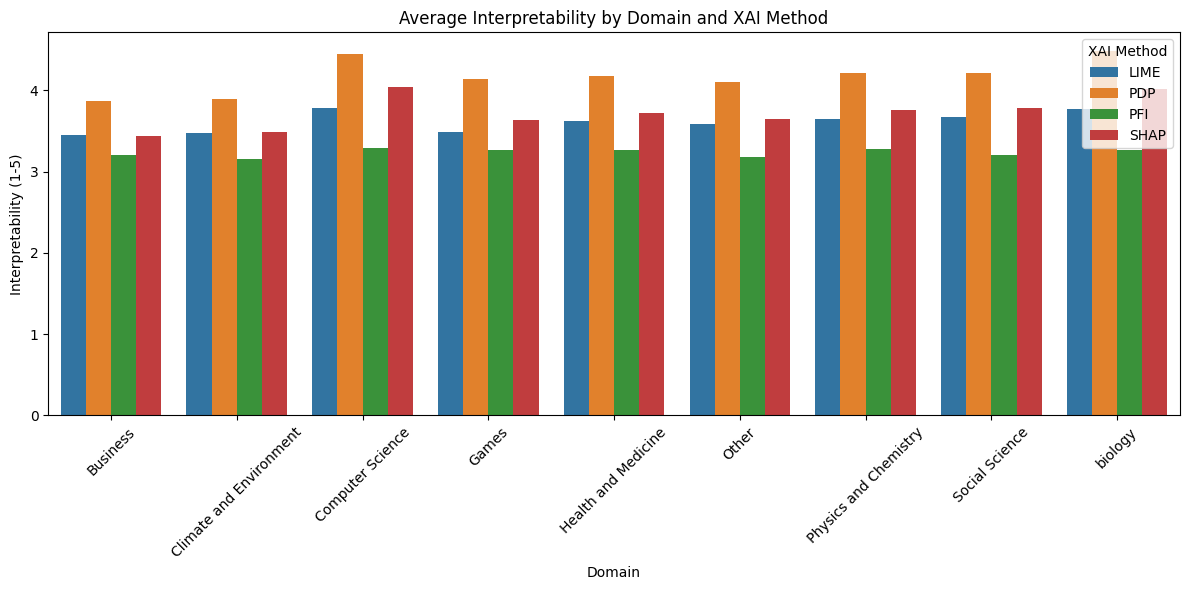}
\caption{Average interpretability (1--5 scale) by domain and XAI method. Higher bars indicate that end-users (virtual personas) found the explanations more understandable and trustworthy.}
\label{fig:avg_interpretability_domain}
\end{figure}

We measure user-perceived \emph{interpretability} by aggregating virual personas ratings (interpretability, understanding, trust). Figure~\ref{fig:avg_interpretability_domain} displays these average interpretability scores by domain. PDP appears to be the highest-scoring method in nearly every domain, indicating that users find its partial dependence approach particularly intuitive or easy to understand. While LIME and SHAP also achieve relatively strong scores in several domains (e.g., business, health and medicine), their performance is slightly outpaced by PDP in most cases. PFI, meanwhile, maintains moderate interpretability levels overall. These results underscore the value of considering multiple XAI methods in practice; while PDP may be a strong default choice in many scenarios, certain domains or user requirements might favor the localized explanations of LIME or the feature-attribution clarity of SHAP.

\subsection{Repository Construction and Domain-Specific Recommendations}
All these data sources are integrated into a unified repository keyed by \emph{dataset\_id}.

When a user uploads a new dataset, our system extracts its characteristics (e.g., feature count, numeric/categorical ratio, missing ratio), then computes similarity to benchmark datasets in the repository using cosine similarity. The top-$k$ similar datasets’ XAI metrics and user ratings are aggregated to estimate a multidimensional XAI score for each method.

\section{Conclusion}
Our experimental results demonstrate that:
\begin{enumerate}
    \item Certain domains (\emph{health and medicine}) contain more datasets, reinforcing the strong emphasis on interpretability in clinical settings.
    \item Quantitative metrics vary significantly by domain and XAI method, suggesting that no single method universally dominates across all contexts.
    \item End-users’ subjective interpretability ratings often diverge from purely technical measures such as fidelity or simplicity, emphasizing the need for user-driven evaluations.
    \item Domain-specific synergy is crucial.
\end{enumerate}

Hence, our approach merges quantitative benchmarking with qualitative persona insights, further refined by domain bonuses from JSON surveys. This fusion significantly advances previous frameworks that relied solely on one dimension of evaluation.

In future work, we plan to expand this approach to other data modalities (e.g., text, images) and refine the synergy model that determines domain bonuses.

\section{Acknowledgement}
The research leading to the results presented in this paper has received funding from the Europeans Union's funded Project FAME under grant agreement no 101092639 and HumAIne under grant agreement no 101120218.

\bibliographystyle{IEEEtran}
\bibliography{references}

\end{document}